%% file: acl_latex.tex
\documentclass[11pt]{article}

\usepackage[preprint]{acl}

\usepackage{times}
\usepackage{latexsym}

\usepackage[T1]{fontenc}

\usepackage[utf8]{inputenc}
\usepackage{xspace}
\usepackage{microtype}

\usepackage{inconsolata}

\usepackage{graphicx}
\usepackage[table]{xcolor}

\usepackage[utf8]{inputenc} 
\usepackage[T1]{fontenc}    
\usepackage{hyperref}       
\usepackage{url}            
\usepackage{booktabs}       
\usepackage{amsfonts}       
\usepackage{nicefrac}       
\usepackage{microtype}      
\usepackage{xcolor}         
\usepackage{amsmath}
\usepackage{graphicx}
\usepackage{tabularx}
\usepackage[most]{tcolorbox}
\tcbuselibrary{listings,skins,breakable}
\definecolor{scablue}{RGB}{21,86,145}

\newsavebox{\trbox}

\usepackage{array}

\newenvironment{transcript}[1]{%
  \begin{tcolorbox}[colback=gray!3, colframe=gray!50, boxrule=0.4pt, arc=2pt,
                    left=3pt, right=3pt, top=2pt, bottom=2pt,
                    fontupper=\scriptsize]%
  \begin{tabular}{@{}p{4.3em}@{\hspace{3pt}}>{\raggedright\arraybackslash}p{\dimexpr\linewidth-4.3em-3pt\relax}@{}}}
  {\end{tabular}\end{tcolorbox}}

\newcommand{\trow}[2]{\textbf{#1} & #2 \\}
\newcommand{\scarow}[2]{\rowcolor{blue!8}\textbf{#1} & #2 \\}

\newcommand{\emphbf}[1]{{\textbf{#1}\xspace}}

\newcommand{\mypara}[1]{\smallskip\noindent\emphbf{#1.}\xspace}

\newtcblisting{promptbox}[1]{
  enhanced, breakable, listing only,
  colback=gray!5, colframe=gray!20,
  fonttitle=\bfseries\sffamily\small,
  coltitle=black,
  title=#1,
  boxrule=0.4pt, arc=1pt,
  listing options={
    basicstyle=\ttfamily\fontsize{7.5pt}{8.6pt}\selectfont,
    breaklines=true,
    breakatwhitespace=false,
    breakindent=0pt,
    breakautoindent=false,
    columns=fullflexible,
    showstringspaces=false,
    keepspaces=true,
    xleftmargin=0pt,
  }
}

\usepackage{xcolor}

\definecolor{IlliniOrange}{RGB}{232, 74, 39}

\definecolor{SBBlue}{RGB}{0, 102, 204}

\title{Symbolic Constraint Agents: Mining Formal Tool-Call Constraints to Prevent Agent Errors}

\author{Jinu Lee\textsuperscript{*} \\
  University of Illinois Urbana-Champaign \\
  \texttt{jinulee2@illinois.edu} \\\And
  Shraddha Barke \\
  Microsoft Research \\
  \texttt{sbarke@microsoft.com} \\}

\begin{document}
\maketitle
\begingroup
\def\thefootnote{*}
\footnotetext{Work done during an internship at Microsoft Research.}
\endgroup

\begin{abstract}
Symbolic constraints (programmatic, deterministic checks for tool calls) enable deterministic verification of LLM agents' tool calls, but existing approaches primarily enforce predefined safety policies rather than tool dependencies in complex tool environments. We explore whether symbolic constraints can be mined from environments and observations beyond predefined policies, and whether they can improve end-to-end task performance. 
We present Symbolic Constraint Agents (SCA), a framework that mines symbolic constraints from API documentation, refines them against execution traces, and checks the agent's tool calls with an SMT solver at runtime. A violation triggers a natural-language warning or, when the valid replacement is uniquely determined, an automatic repair. On AppWorld, refinement raises error-detection F1 from 37.0 to 58.3 on GPT-5.5 traces and from 39.3 to 61.9 on Qwen3.5-27B traces. At runtime, repair raises GPT-5.4-mini's task success by 5.4 points under a 40-call budget and 9.8 points when environment errors are hidden, while removing 59--87\% of failed calls across models, and transfers to $\tau^2$-bench with gains of up to 5.6 points. These findings show that automatically acquired symbolic constraints can improve tool-calling reliability and task completion, particularly when failures are costly or difficult to diagnose.
\end{abstract}

\input{intro}
\input{related-work}
\input{symbolic-agents}
\input{eval}
\input{results}
\input{conclusion}

\newpage
\section*{Limitations}

Constraints are built once with a strong proprietary model (GPT-5.5) and refined on execution traces of training tasks, which requires access to such traces and incurs a one-time cost.
Our constraints check each tool call against the call history, so they detect errors that are visible in that history: wrong or miscopied arguments, calls in the wrong state, missing prerequisites.
They cannot detect failures that depend on hidden environment state (e.g., a credit limit the agent has not queried), and they cannot detect omissions, where the agent never makes a necessary call.
This is why recall is much lower than precision (Table~\ref{tab:error-detection}), and why constraints help little for agents whose failures are mostly of these kinds, such as Qwen3.5-9B.

Improvements in task success are largest when tool errors are costly or invisible and when the agent makes many call-level errors.
In the standard AppWorld setting, success gains are small, and a strong model that rarely makes these errors (Claude Sonnet-5) benefits little.
Warnings can also mislead: a false alarm may cause an agent to retry a call that was valid.

\section*{Ethical Considerations}
Symbolic constraints make tool-use agents more reliable, but they are not a safety guarantee: mined rules can be wrong or too strict, and a too-strict rule can flag or alter a legitimate action. More reliable agents can also be used to automate harmful actions, which our checker does not prevent. All experiments run in simulated environments (AppWorld, $\tau^2$-bench) with synthetic users and no personal data. Constraint mining is a one-time cost per environment, and repair reduces failed tool calls and, for the weaker models, LLM calls and tokens per task (Table~\ref{tab:cost}).

\bibliography{custom}

\appendix

\section{Formal Definition of SCA}
\label{app:z3}

This section describes the formal logic machinery for detecting and repairing constraint-violating tool calls under SCA.

\subsection{Sorts and Predicates}

SCA introduces four sorts: a unique ID for each tool call, an enumeration of possible tool names and field names, and parameter/return values. Every concrete parameter or return value observed in a trace is lifted into exactly one of the five constructors.
\begingroup\small
\begin{align*}
  \mathsf{CallID}    &: \mathsf{String} \\
  \mathsf{ToolName}  &: \text{enumerated sort} \\
  \mathsf{FieldName} &: \text{enumerated sort} \\
  \mathsf{Value}     &::= \mathtt{v\_str}(\mathsf{String}) \mid \mathtt{v\_int}(\mathbb{Z}) \mid \mathtt{v\_real}(\mathbb{R}) \\
                     &\phantom{::=} \mid \mathtt{v\_bool}(\mathbb{B}) \mid \mathtt{v\_null}
\end{align*}
\endgroup
Using these sorts, we define four predicates (uninterpreted functions) to express tool-call events.
\begingroup\small
\begin{align*}
  \mathsf{call}    &: \mathbb{N}_0 \times \mathsf{CallID} \times \mathsf{ToolName} \to \mathbb{B} \\
  \mathsf{params}  &: \mathsf{CallID} \times \mathsf{FieldName} \times \mathsf{Value} \to \mathbb{B} \\
  \mathsf{returns} &: \mathsf{CallID} \times \mathsf{FieldName} \times \mathsf{Value} \to \mathbb{B} \\
  \mathsf{finish}  &: \mathbb{N}_0 \to \mathbb{B}
\end{align*}
\endgroup
Here $\mathsf{call}(i,c,T)$ states that step $i$ is call $c$ of tool $T$; $\mathsf{params}(c,f,v)$ states that call $c$ was invoked with parameter $f = v$; $\mathsf{returns}(c,f,v)$ states that call $c$ returned field $f = v$; and $\mathsf{finish}(i)$ states that step $i$ is the trace's terminal action.

\subsection{Axioms}

\paragraph{Closed-world assumptions.} Simply asserting the uninterpreted functions as ground facts would leave models free to also satisfy the predicate on tuples never observed. In other words, SMT solvers may invent non-existent values or allow duplicate values registered for the same tool call. Hence, we add closed-world axioms.

Specifically, let $F_p$ be the domain as the set of all observed tuples for each uninterpreted predicate $p\in\{\mathsf{call}, \mathsf{params}, \mathsf{returns}\}$, and $F_{\mathsf{id}}$ be the set of already instantiated $\mathsf{CallID}$s. At turn $i_t$, we define the axioms as:
\begingroup\small
\begin{align*}
  &\textbf{cwa\_call:} \\
  &\quad \forall i, c, T.\;
    \big(\mathsf{call}(i,c,T) \wedge i \leq i_t\big) \rightarrow {} \\
  &\qquad \textstyle\bigvee_{(i_k,c_k,T_k)\in F_{\mathsf{call}}}
      \big(i{=}i_k \wedge c{=}c_k \wedge T{=}T_k\big)
  \\[5pt]
  &\textbf{cwa\_params:} \\
  &\quad \forall c,f,v.\;
    \big(\mathsf{params}(c,f,v) \wedge \textstyle\bigvee_{c_k \in F_{\mathsf{id}}} c{=}c_k\big)
    \rightarrow {} \\
  &\qquad \textstyle\bigvee_{(c_k,f_k,v_k)\in F_{\mathsf{params}}}
      \big(c{=}c_k \wedge f{=}f_k \wedge v{=}v_k\big)
  \\[5pt]
  &\textbf{cwa\_returns:} \\
  &\quad \forall c,f,v.\;
    \big(\mathsf{returns}(c,f,v) \wedge \textstyle\bigvee_{c_k \in F_{\mathsf{id}}} c{=}c_k\big)
    \rightarrow {} \\
  &\qquad \textstyle\bigvee_{(c_k,f_k,v_k)\in F_{\mathsf{returns}}}
      \big(c{=}c_k \wedge f{=}f_k \wedge v{=}v_k\big)
\end{align*}
\endgroup

\paragraph{Structural axioms.} We further introduce more axioms to make sure that \textit{future} traces are also valid.
\begingroup\small
\begin{align*}
  &\textbf{call\_id\_uniq:} \\
  &\quad \forall c,i,j,T_1,T_2.\;
    \mathsf{call}(i, c, T_1)\wedge\mathsf{call}(j, c, T_2) \\
  &\qquad \rightarrow\; (i{=}j \wedge T_1{=}T_2)
  \\[5pt]
  &\textbf{one\_call\_per\_ts\_exist:} \\
  &\quad \forall i.\;
    \mathsf{finish}(i)\vee\exists c,T.\;\mathsf{call}(i,c,T)
  \\[5pt]
  &\textbf{one\_call\_per\_ts\_uniq:} \\
  &\quad \forall i,c_1,c_2,T_1,T_2.\;
    \mathsf{call}(i,c_1,T_1) \wedge \mathsf{call}(i,c_2,T_2) \\
  &\qquad \rightarrow\; (c_1{=}c_2 \wedge T_1{=}T_2)
  \\[5pt]
  &\textbf{param\_field\_exist}[T,f]\textbf{:} \\
  &\quad \forall i,c.\;
    \mathsf{call}(i, c, T) \rightarrow \exists v.\;\mathsf{params}(c, f, v)
  \\[5pt]
  &\textbf{param\_field\_uniq:} \\
  &\quad \forall c,f,v_1,v_2.\;
    \mathsf{params}(c,f,v_1) \wedge \mathsf{params}(c,f,v_2) \\
  &\qquad \rightarrow\; v_1{=}v_2
  \\[5pt]
  &\textbf{param\_call\_id\_exist:} \\
  &\quad \forall c,f,v.\;
    \mathsf{params}(c,f,v) \rightarrow \exists i,T.\;\mathsf{call}(i,c,T)
  \\[5pt]
  &\textbf{return\_field\_exist}[T,f]\textbf{:} \\
  &\quad \forall i,c.\;
    \mathsf{call}(i, c, T) \rightarrow \exists v.\;\mathsf{returns}(c, f, v)
  \\[5pt]
  &\textbf{return\_field\_uniq:} \\
  &\quad \forall c,f,v_1,v_2.\;
    \mathsf{returns}(c,f,v_1) \wedge \mathsf{returns}(c,f,v_2) \\
  &\qquad \rightarrow\; v_1{=}v_2
  \\[5pt]
  &\textbf{return\_call\_id\_exist:} \\
  &\quad \forall c,f,v.\;
    \mathsf{returns}(c,f,v) \rightarrow \exists i,T.\;\mathsf{call}(i,c,T)
  \\[5pt]
  &\textbf{no\_calls\_after\_finish:} \\
  &\quad \forall i,j,c,T.\;
    \mathsf{finish}(i) \wedge j{\geq}i \rightarrow \neg\,\mathsf{call}(j,c,T) \\
  &\quad \forall i,j.\;
    \mathsf{finish}(i) \wedge j{>}i \rightarrow \neg\,\mathsf{finish}(j)
\end{align*}
\endgroup

Two parameterized existence axioms, \textbf{param\_field\_exist} and \textbf{return\_field\_exist}, are instantiated with all combinations of tool and its corresponding required fields.

In reality, instantiating rules for all tools and their parameters/return values incurs a significant runtime increase in Z3. To reduce the delay, we limit the universe of tools to those exposed to the agent via tool search.

\subsection{Realizability Query}
\label{app:query}

Given a trace and the current catalog of mined constraints
$\{\varphi_1,\dots,\varphi_n\}$ (each $\varphi_i$ a closed first-order
formula over $\mathsf{call}/\mathsf{params}/\mathsf{returns}$ in the
vocabulary above), the checker asks, under the closed-world and structural axioms:
\[
  \mathrm{SAT}\Big[\; \bigwedge_{k} \textbf{ax}_k \;\wedge\; \bigwedge_{i} \varphi_i \;\Big].
\]
Each axiom and each constraint is asserted as a separately-tracked
assumption. An $\mathrm{UNSAT}$ verdict yields an UNSAT core, a subset of axioms and constraints that are jointly unsatisfiable, which is
then used to attribute a violation to a specific call in the trace. This information is used for the constraint refinement (Section~\ref{subsec:constgen}) and warning (Section~\ref{subsec:application}).
\section{Implementation details}
\label{app:implementation}

SCA is implemented as an agent in OpenHands~\citep{wang2025openhands}. It subclasses the CodeAct agent~\citep{wang2024executable}, so the underlying LLM agent, its prompts, and its tool handling are unchanged. The environment's tools are served to the agent over a Model Context Protocol (MCP) server.
SCA operates by intercepting each MCP observation before it reaches the LLM. It observes the called tool, its arguments, and response, and appends them to a running set of trace facts which are then passed to the constraint checker.

Constraint checking uses Z3 (v4.16) through its Python API. Tool and field names are Z3 enumeration sorts, and argument and return values are an algebraic datatype covering strings, integers, reals, booleans and null; lists and maps are unrolled into their primitive leaves.
For each check, the trace facts are substituted into the uninterpreted predicates \texttt{call}, \texttt{params}, \texttt{returns} and \texttt{finish} of the scoped rules. Z3 then checks the result with tracked assertions, and the unsat core identifies the violated rules.
To keep checks fast, SCA lifts trace facts into Z3 terms incrementally and caches them across steps. It asserts only the rules scoped to the called tool, and each check has a 5\,s timeout.
A warning is appended in place to the text of the MCP observation.
A repair runs as a dry-run check on the proposed MCP action before dispatch, using a snapshot of the trace and cache that is rolled back afterwards. When the repair is unique, SCA rewrites the action's arguments in place, so the environment executes only the corrected call.
Repairs are computed with Z3's \texttt{Optimize} (MaxSAT with soft constraints on the original values), followed by a second solver call that checks uniqueness.
Every check, warning, and repair is logged with its verdict and the rules involved.
\section{Prompts}
\label{app:prompts}

We reproduce the prompts of the initializer and refiner agents below. Bracketed segments in capitals are placeholders filled in at runtime.

\begin{promptbox}{Initializer agent prompt}
Final-answer format (no markdown fences, no prose, no tool calls): a single JSON object with exactly these two keys:

- "join": list[{"rid": int, "rationale": str}]
  Add the CURRENT TARGET TOOL to the `applies_to` set of the existing rule #rid shown under EXISTING RULES. Use this -- instead of writing a near-duplicate -- when the target tool's same-named parameter obeys the SAME invariant as that rule (same provenance source, same type or value-domain constraint). Joining reuses the rule's z3 verbatim; do NOT restate it.

- "add": list[{"nl_constraint": str, "z3_constraint": str, "source_tools": list[str], "rationale": str}]
  Author a brand-new rule that (for now) applies only to the current target tool. The `z3_constraint` MUST name the target call with the `call(i1, c1, __TARGET__)` placeholder. `source_tools` lists any concrete tools a provenance rule reads from (omit / [] otherwise).

Use [] for either key you do not need. Return {"join": [], "add": []} if no high-quality constraint applies to this tool. Every assistant turn must be either (a) one or more search-tool calls, or (b) the final JSON object.

========================================

TARGET TOOL DEFINITION:
[TARGET TOOL DEFINITION]

EXISTING RULES that constrain a parameter whose name matches one of this tool's parameters (candidates to JOIN):
[EXISTING RULES, or: (no existing rules constrain a parameter of this tool yet)]

For EACH existing rule above, decide whether this target tool's same-named parameter obeys the SAME invariant (same provenance source, same type / value-domain constraint). If so, `join` that rule (rid) -- do NOT write a new rule. If this tool's parameter follows a DIFFERENT invariant, or a needed invariant is not covered by any rule above, `add` a new rule using the `call(i1, c1, __TARGET__)` placeholder for the target call.

You do NOT need to search for sibling tools that share a parameter -- the catalog grows incrementally as each tool is processed. Use the search-tool functions only to inspect a provenance SOURCE tool you must reference. Make sure each constraint triggers on incorrect traces but never invalidates correct ones. Data-provenance rules must be high quality: only require a parameter to come from a prior return when the exact source tool and return field are supported by inspected schemas; if the data could legitimately come from arbitrary different tools, do not generate a constraint. Do NOT add constraints about the tool's own return values unless an obvious invariant exists.
\end{promptbox}

\begin{promptbox}{Refiner agent prompt}
Respond with a single JSON object (no markdown fences, no prose) with exactly these keys:

- "remove": list[{"rid": int, "rationale": str}]
  Delete rule #rid from the catalog. `rationale` must concisely justify the removal (e.g. "spuriously fired: trace shows X is allowed without Y", "subsumed by #42", "wrong scope: applies to tool A not B").

- "edit": list[{"rid": int, "nl_constraint": str, "z3_constraint": str, "rationale": str}]
  Replace rule #rid with the given new nl/z3 pair. `rationale` must explain *why* the rule is being changed and what specifically in the trace or current rule motivated it (e.g. "tightened scope to only apps_phone:show_account", "fixed off-by-one on price comparison"). The new nl_constraint / z3_constraint MUST be a generic, reusable invariant -- not phrased in terms of "this task" / "this prompt" / prompt-specific literals (see GENERALITY REQUIREMENT in the user message).

- "add": list[{"tool_name": str, "nl_constraint": str, "z3_constraint": str, "rationale": str}]
  Append a brand-new rule under the named tool group. The tool_name must exactly match a name in the RELEVANT TOOLS or CURRENT RULE CATALOG. `rationale` must explain what violation in the trace this new rule would have prevented and why no existing rule catches it. The nl_constraint / z3_constraint MUST be a generic invariant that will fire on ANY future trace exhibiting the same violation pattern -- not a rule scoped to this specific user prompt or its literal values (see GENERALITY REQUIREMENT in the user message).

Use [] for any operation you do not need. Do not include any other top-level keys. Every op MUST include a non-empty `rationale` string; ops missing a rationale will be dropped.

You have access to a `suggest_edit` tool. Use it to TEST your proposed changes BEFORE committing them:

1. For each unsat-core rule, examine the specific tool call(s) that triggered it. Check the exact parameter values against the return values of prior calls.
2. Also examine tool calls that FAILED (returned error messages). If a call failed due to an API contract violation (wrong type, missing required field, invalid value) and no existing rule captures that invariant, propose ADDING a new rule.
3. If you determine a rule is genuinely wrong, remove it. If you want to change a rule, remove the old rule and add a replacement rule. If you want to add a new rule, include an add operation with `applies_to`, `nl_constraint`, `z3_constraint`, and `rationale`.
4. Inspect the returned SAT verdict, final UNSAT core, and step-wise UNSAT cores. Use `current_call_core_rids` to identify which proposed catalog rules are still blamed on each specific tool call; `full_unsat_core_rids` may include stale cumulative poison from earlier calls.
5. If you cannot find a valid reason to change any rule AND no uncaptured violations exist, output an empty ops object: {"remove": [], "add": []}. This means the constraint catalog is already correct for this trace.

========================================

USER PROMPT:
[ORIGINAL TASK INSTRUCTION]

EXECUTION TRACE (chronological tool calls):
[TRACE + PER-STEP CONSTRAINT CHECK RESULTS]

IMPORTANT -- GENERALITY REQUIREMENT: this catalog is GLOBAL and is enforced against ALL FUTURE traces from ALL FUTURE user prompts, not only the one above. Every `add` you propose MUST therefore be phrased as a generic, reusable invariant about the tools / API surface -- not a rule scoped to this specific trace's user prompt.

IMPORTANT -- EXHAUSTIVE SOURCE ENUMERATION: when writing a constraint that says 'the value X used by tool A must have been produced earlier by some tool', you MUST:
(1) Use the search tool EXHAUSTIVELY to discover ALL tools whose response schemas can produce that value. A common mistake is listing only the obvious sources while missing others that also return the same field in their response.
(2) Remember that a user's task instruction can supply values directly (e.g. a file path, a name, an ID). A constraint that demands a value MUST come from a prior tool return will spuriously fire when the user simply told the agent the value. Only write provenance constraints for values that genuinely cannot be known without a prior API call.
\end{promptbox}

\begin{table}[h]
  \centering
  \footnotesize
  \setlength{\tabcolsep}{2pt}
  \begin{tabular}{llrrrrr}
    \toprule
    Benchmark & Catalog & Rules & Arg. & Hist. & Outc. & Tools \\
    \midrule
    AppWorld          & refined            & 295 & 239 & 47 & 9 & 457 \\
    \midrule
    $\tau^2$ airline  & policy, initial    & 51  & 36  & 15 & 0 & 13 \\
                      & policy, refined    & 51  & 37  & 14 & 0 & 13 \\
                      & no policy, initial & 44  & 41  & 3  & 0 & 13 \\
                      & no policy, refined & 44  & 41  & 3  & 0 & 13 \\
    \midrule
    $\tau^2$ retail   & policy, initial    & 69  & 36  & 33 & 0 & 16 \\
                      & policy, refined    & 63  & 36  & 26 & 1 & 16 \\
                      & no policy, initial & 47  & 32  & 15 & 0 & 15 \\
                      & no policy, refined & 47  & 31  & 16 & 0 & 15 \\
    \bottomrule
  \end{tabular}
  \caption{Constraint catalogs: number of rules by type (argument, history, outcome) and number of tools they apply to. AppWorld rules cover 9 apps.
  On $\tau^2$-bench, refinement on the training split replaced 10 of 51 airline and 18 of 69 retail policy rules (adding 10 and 12), and 4 of 44 airline and 6 of 47 retail no-policy rules (adding 4 and 6).}
  \label{tab:catalog-stats}
\end{table}

\begin{table*}[h]
  \centering
  \small
  \begin{tabular}{p{\textwidth}}
    \toprule
    \begin{minipage}[t]{\linewidth}\raggedright
      \textbf{AppWorld, argument rule} \quad applies to \texttt{amazon\_\_add\_product\_to\_cart (+4 more)}\par
      \textit{If the quantity parameter is provided, it must be an integer greater than or equal to 1.}\par\vspace{2pt}
      {\scriptsize\texttt{ForAll([i1, c1, v1], Implies(And(call(i1, c1, \_\_TARGET\_\_), params(c1, quantity\_f, v1)), And(Value.is\_v\_int(v1), Value.v\_int\_get(v1) >= 1)))}\par}\vspace{2pt}
    \end{minipage} \\
    \midrule
    \begin{minipage}[t]{\linewidth}\raggedright
      \textbf{AppWorld, history rule} \quad applies to \texttt{gmail\_\_create\_draft (+41 more)}\par
      \textit{The access\_token parameter must be the access\_token returned by a prior gmail login call.}\par\vspace{2pt}
      {\scriptsize\texttt{ForAll([i1, c1, v1], Implies(And(call(i1, c1, \_\_TARGET\_\_), params(c1, access\_token\_f, v1)), Exists([i2, c2], And(i2 < i1, And(call(i2, c2, appworld\_gmail\_\_login), rfield(c2, access\_token\_f, v1))))))}\par}\vspace{2pt}
    \end{minipage} \\
    \midrule
    \begin{minipage}[t]{\linewidth}\raggedright
      \textbf{AppWorld, outcome rule} \quad applies to \texttt{amazon\_\_place\_order}\par
      \textit{An Amazon place\_order call must not be attempted while the cart has an invalid promo code applied; it must not return the invalid-promo-code failure message.}\par\vspace{2pt}
      {\scriptsize\texttt{ForAll([i1, c1, v1], Implies(And(call(i1, c1, \_\_TARGET\_\_), rfield(c1, message\_f, v1), Value.is\_v\_str(v1)), Value.v\_str\_get(v1) != StringVal("The promo code applied to your cart is no longer valid.")))}\par}\vspace{2pt}
    \end{minipage} \\
    \midrule
    \begin{minipage}[t]{\linewidth}\raggedright
      \textbf{$\tau^2$ retail, policy, history rule} \quad applies to \texttt{calculate}\par
      \textit{Before any calculate tool call, the user must already have been authenticated by a prior lookup using email or name plus zip code.}\par\vspace{2pt}
      {\scriptsize\texttt{ForAll([i1, c1], Implies(call(i1, c1, calculate), Exists([i2, c2], And(i2 < i1, Or(call(i2, c2, find\_user\_id\_by\_email), call(i2, c2, find\_user\_id\_by\_name\_zip))))))}\par}\vspace{2pt}
    \end{minipage} \\
    \midrule
    \begin{minipage}[t]{\linewidth}\raggedright
      \textbf{$\tau^2$ airline, no policy, history rule} \quad applies to \texttt{update\_reservation\_passengers}\par
      \textit{Before updating passengers for a reservation, the same reservation\_id must have been looked up earlier with get\_reservation\_details.}\par\vspace{2pt}
      {\scriptsize\texttt{ForAll([i1, c1, v1], Implies(And(call(i1, c1, update\_reservation\_passengers), params(c1, reservation\_id\_f, v1), Value.is\_v\_str(v1)), Exists([i2, c2], And(i2 < i1, call(i2, c2, get\_reservation\_details), params(c2, reservation\_id\_f, v1)))))}\par}\vspace{2pt}
    \end{minipage} \\
    \bottomrule
  \end{tabular}
 \caption{Example constraints, copied verbatim from the catalogs used in our experiments: the natural-language rule (italics) and its SMT formula. The outcome rule is the one that fires in the warning example (Figure~\ref{fig:warning-example}); nearly every repair in our runs enforces an access-token history rule like the gmail one shown.}
  \label{tab:catalog-examples}
\end{table*} 

\section{Constraint Catalogs}
\label{app:catalogs}

Each constraint pairs a natural-language rule with an SMT formula over the call history.
The formula quantifies over calls \texttt{call(i, c, tool)} (position \texttt{i}, call \texttt{c}), their parameters \texttt{params(c, field, value)}, and the fields of their responses \texttt{rfield(c, field, value)}, the $\operatorname{returns}$ predicate of Section~\ref{subsec:formulation}; \texttt{\_\_TARGET\_\_} stands for each tool the rule applies to.
We group rules by what their formula refers to:
\emph{argument} rules check only the arguments of the call itself (e.g., value ranges and formats);
\emph{history} rules relate the call to other calls in the history, e.g., that an access token was returned by a prior login to the same app;
\emph{outcome} rules read the response of the call itself, e.g., that an order did not fail because the payment card has expired, and can therefore only be checked after the call returns (warning mode).
Table~\ref{tab:catalog-stats} summarizes the catalogs used in our experiments, and Table~\ref{tab:catalog-examples} shows one example from each type.


\section{Detailed Audit on Constraints}
\label{app:rule-audit}

We audit every rule of the five catalogs used in our experiments (500 rules; Table~\ref{tab:catalog-stats}).
Three LLM judges from two model families (GPT-5.6 Sol, GPT-6 Astra, and Claude Opus 5.5) label each rule independently.
Each judge sees the natural-language rule, its corresponding Z3 formula, the definitions of the tools mentioned in the formula, and the domain policy (for $\tau$-bench).
The judge answers two questions:
\emph{faithful} -- does the formula express the same condition as the natural-language rule? -- and
\emph{validity} -- is the rule \emph{correct} (violating calls would fail or break a domain rule, and valid calls satisfy it), \emph{too strict} (it also rejects some legitimate calls), \emph{wrong}, or \emph{unsure}.
We take the majority label; there was a majority for every rule.
All three judges agree on 87\% of rules for faithfulness and 78\% for validity.

\begin{table}[t]
  \centering
  \footnotesize
  \setlength{\tabcolsep}{4pt}
  \begin{tabular}{lrrr}
    \toprule
    Catalog & Rules \# & Faithful\% & Validity\% \\
    \midrule
    AppWorld                     & 295 & 94 & 89  \\
    $\tau^2$ airline, policy     & 51  & 84 & 96  \\
    $\tau^2$ airline, no policy  & 44  & 91 & 100 \\
    $\tau^2$ retail, policy      & 63  & 81 & 95  \\
    $\tau^2$ retail, no policy   & 47  & 100 & 94 \\
    \midrule
    Argument rules               & 384 & 93 & 99  \\
    History rules                & 106 & 89 & 67  \\
    Outcome rules                & 10  & 70 & 100 \\
    \midrule
    All                          & 500 & 92 & 92  \\
    \bottomrule
  \end{tabular}
  \caption{Rule audit by three LLM judges (majority vote). No rule was judged \emph{wrong} or \emph{unsure} for validity; every rule that was not \emph{correct} (1-validity) is \emph{too strict}.}
  \label{tab:rule-audit}
\end{table}

To validate the LLM judge evaluations, one author labeled 50 rules without seeing the three LLM judges' labels: 35 rules flagged by 2 or more judges and 15 rules accepted by all three.
The author agreed with the judges on 94\% of rules for faithfulness and 90\% for validity, showing strong human-LLM agreement.
There were 5 cases where the author judged the rule as valid, but the judges did not; for example, the judges called ``\texttt{task\_comment\_id} must come from an earlier call that created or listed comments'' too strict because an ID could come from the task, but in AppWorld such IDs can only come from those calls.

Table~\ref{tab:rule-audit} summarizes the results.
The ratios of faithful and valid rules are both measured around 92\%. While argument rules and outcome rules are almost always correct (99\% and 100\%, respectively), history rules remain only 67\% correct. Most of the errors concentrate on rules that assume an identifier (an address, a payment card, a comment) to be returned by an earlier call, since a legitimate value could also come from the task description (Table~\ref{tab:rule-audit-examples}, second example).


\begin{table*}[t]
  \centering
  \small
  \begin{tabular}{p{0.97\textwidth}}
    \toprule
    \begin{minipage}[t]{\linewidth}\raggedright
      \textbf{Unfaithful (3/3 judges), correct (3/3)} \quad AppWorld, argument rule, applies to \texttt{amazon\_\_signup} (+15 more)\par
      \textit{Every signup call must provide a first\_name parameter that is a non-empty string.}\par\vspace{2pt}
      {\scriptsize\texttt{ForAll([i1, c1, v1], Implies(And(call(i1, c1, \_\_TARGET\_\_), params(c1, first\_name\_f, v1)), And(Value.is\_v\_str(v1), Length(Value.v\_str\_get(v1)) >= 1)))}\par}\vspace{2pt}
      Judge (Opus): ``The formula only checks first\_name when it is supplied and never requires it to be present, while the rule says every signup call must provide it.''\par\vspace{2pt}
    \end{minipage} \\
    \midrule
    \begin{minipage}[t]{\linewidth}\raggedright
      \textbf{Faithful (3/3), too strict (3/3)} \quad AppWorld, history rule, applies to \texttt{amazon\_\_place\_order} (+2 more)\par
      \textit{The address\_id parameter must be an address\_id returned by a prior Amazon add\_address or show\_addresses call.}\par\vspace{2pt}
      {\scriptsize\texttt{ForAll([i1, c1, v1], Implies(And(call(i1, c1, \_\_TARGET\_\_), params(c1, address\_id\_f, v1)), Exists([i2, c2], And(i2 < i1, Or(call(i2, c2, appworld\_amazon\_\_add\_address), call(i2, c2, appworld\_amazon\_\_show\_addresses)), rfield(c2, address\_id\_f, v1)))))}\par}\vspace{2pt}
      Judge (Opus): ``Address IDs could legitimately come from other sources, such as order details or IDs given in the task, so valid calls without a prior add\_address or show\_addresses would be rejected.''\par\vspace{2pt}
    \end{minipage} \\
    \midrule
    \begin{minipage}[t]{\linewidth}\raggedright
      \textbf{Faithful (3/3), too strict (3/3)} \quad $\tau^2$ retail, no policy, history rule, applies to \texttt{cancel\_pending\_order}\par
      \textit{An order may not be cancelled more than once, so no earlier cancel\_pending\_order call may have used the same order\_id.}\par\vspace{2pt}
      {\scriptsize\texttt{ForAll([i1, c1, v1], Implies(And(call(i1, c1, cancel\_pending\_order), params(c1, order\_id\_f, v1)), Not(Exists([i2, c2], And(i2 < i1, call(i2, c2, cancel\_pending\_order), params(c2, order\_id\_f, v1))))))}\par}\vspace{2pt}
      Judge (Opus): ``The formula does not check whether the earlier call succeeded, so a legitimate retry after a failed attempt (e.g., one rejected for an invalid reason string) would wrongly be flagged.''\par\vspace{2pt}
    \end{minipage} \\
    \midrule
    \begin{minipage}[t]{\linewidth}\raggedright
      \textbf{Faithful (3/3), correct (2/3)} \quad AppWorld, history rule, applies to \texttt{gmail\_\_create\_draft} (+41 more); the rule behind most repairs\par
      \textit{The access\_token parameter must be the access\_token returned by a prior gmail login call.}\par\vspace{2pt}
      Judge (Opus, correct): ``The documentation says the access\_token is `obtained from gmail app login', so a valid call needs a token returned by a prior gmail login.''\par
      Judge (Astra, too strict): ``A valid Gmail access token obtained before the recorded call history can authenticate a legitimate call without a prior login call in that history.''\par\vspace{2pt}
    \end{minipage} \\
    \bottomrule
  \end{tabular}
  \caption{Examples from the rule audit: the judges' labels, the rule (italics) and its formula, and the judges' reasons, quoted verbatim.}
  \label{tab:rule-audit-examples}
\end{table*}
\section{Additional Experiments}

\subsection{Qwen3.5-27B on AppWorld}
\label{app:qwen27b}

Table~\ref{tab:qwen27b} reports the performance of Qwen3.5-27B (single run) on the three AppWorld settings (Standard, 40-call budget, and silent error; Table~\ref{tab:main}).
The performance pattern matches the other models: repair gains most under silent errors (+3.9), little in the standard setting (+1.2), and nothing under the 40-call budget ($-$1.2), while warnings help slightly under budget and silent errors.
The number of failed tool calls per task drops in every cell, by up to 29\% with repair in the standard setting and 42\% with warnings under silent errors.
Cost (Table~\ref{tab:qwen27b-cost}) follows the pattern of GPT-5.4-mini (Table~\ref{tab:cost}): in the standard setting, repair cuts tool calls by 17\%, LLM calls by 20\% and prompt tokens by 16\%. Under the 40-call budget and silent errors, the cost is essentially flat.
\begin{table*}[t]
  \centering
  \small
  \setlength{\tabcolsep}{2pt}
  \newcommand{\ch}[3]{#1\,$\to$\,#2 {\scriptsize(#3)}}
  \begin{tabular}{llccc|cc}
    \toprule
    & & \multicolumn{3}{c|}{Task success (\%)} & \multicolumn{2}{c}{Failed calls per task} \\
    \cmidrule(lr){3-5} \cmidrule(lr){6-7}
    Benchmark (Setting) & Agent model & No constraints & + Warning & + Repair & + Warning & + Repair \\
    \midrule
    AppWorld (Standard)       & Qwen3.5-27B & 49.9 & 49.4 & 51.1 & \ch{6.9}{6.6}{$-$5}  & \ch{6.9}{5.0}{$-$29} \\
    AppWorld (40-call budget) & Qwen3.5-27B & 44.6 & 45.1 & 43.4 & \ch{1.0}{0.8}{$-$18} & \ch{1.0}{0.8}{$-$22} \\
    AppWorld (Silent errors)  & Qwen3.5-27B & 48.4 & 50.1 & 52.3 & \ch{2.3}{1.3}{$-$42} & \ch{2.3}{1.8}{$-$23} \\
    \bottomrule
  \end{tabular}
  \caption{Task success and failed tool calls for Qwen3.5-27B on AppWorld (417 test tasks), one run per cell.
  Same catalog, settings, and agent configurations as Table~\ref{tab:main}.
  Failed calls: tool calls per task rejected by the environment (under silent errors: answered with an empty response), without constraints $\to$ with the given mode (change in \%).}
  \label{tab:qwen27b}
\end{table*}

\begin{table}[t]
  \centering
  \footnotesize
  \setlength{\tabcolsep}{1pt}
  \newcommand{\cp}[3]{#1$\to$#2\,{\scriptsize(#3)}}
  \newcommand{\cm}[3]{\cellcolor{blue!12}#1$\to$#2\,{\scriptsize($-$#3)}}
  \newcommand{\chh}[3]{\cellcolor{blue!30}#1$\to$#2\,{\scriptsize($-$#3)}}
  \begin{tabular}{l ccc}
    \toprule
    Per task     & Standard            & 40-call             & Silent               \\
    \midrule
    \multicolumn{4}{l}{\textit{Qwen3.5-27B}} \\
    Tool calls   & \cm{76}{64}{17}     & \cp{30}{30}{0}      & \cp{34}{34}{+1}      \\
    LLM calls    & \cm{66}{52}{20}     & \cp{20}{20}{0}      & \cp{23}{22}{$-$4}    \\
    Prompt (k)   & \cm{1670}{1399}{16} & \cp{440}{440}{0}    & \cp{553}{523}{$-$5}  \\
    Compl.\ (k)  & \cp{16.4}{16.3}{0}  & \cp{5.1}{5.3}{+4}   & \cp{6.2}{6.2}{$-$1}  \\
    \bottomrule
  \end{tabular}
  \caption{Cost per AppWorld task for Qwen3.5-27B without constraints $\to$ with repair (change in \%), one run per cell.
  \emph{Prompt}/\emph{Compl.}: prompt and completion tokens, in thousands.
  Shading: savings more than 10\%. Same runs as Table~\ref{tab:qwen27b}.}
  \label{tab:qwen27b-cost}
\end{table}

\subsection{Ablation}
\label{sec:ablation}
In Section \ref{subsec:error-detection}, we show that refinement is essential for improving the error detection performance. We next examine whether refinement is necessary for end-to-end performance. In the warning mode on AppWorld, we compare agents using the catalog produced by initialization alone against agents using the refined catalog. As Table~\ref{tab:ablation} shows, removing refinement lowers task success by 3.5 points for GPT-5.5 and 2.4 points for Qwen3.5-27B. Inspecting the trajectories, we find that agents given spurious warnings often retry, and sometimes loop on, tool calls that were in fact valid: the low precision of the unrefined catalog (Table~\ref{tab:error-detection}) propagates directly into lower task success.

\begin{table}[htbp]
  \centering
  \begin{tabular}{lccc}
    \toprule
                 & \multicolumn{3}{c}{Task success (\%)} \\
    \cmidrule(r){2-4}
    Agent model  & Init+Refine & Init only & $\Delta$ \\
    \midrule
    GPT-5.5      & \textbf{89.3} & 85.9 & $-3.5$ \\
    Qwen3.5-27B  & \textbf{59.7} & 57.3 & $-2.4$ \\
    \bottomrule
  \end{tabular}
  \caption{Ablation of refinement. AppWorld task success (\%) in the warning mode with the catalog refined with agent traces (Init+Refine) and without (Init only); $\Delta$ is the change from removing refinement.}
  \label{tab:ablation}
\end{table}

\subsection{Solver Experiments}
\label{sec:solver}

We measure the Z3 solver's throughput on every tool call of the no-constraint runs, using the full catalogs (AppWorld: four models, 417 tasks; $\tau^2$-bench: three models).

\begin{table}[!t]
\centering
\small
\setlength{\tabcolsep}{3pt}
\begin{tabular}{llrrrr}
\toprule
Catalog & Model & Median & p95 & Time- & Share of \\
 & & (ms) & (ms) & outs & ep.\ time \\
\midrule
AppWorld & GPT-5.4-mini & 351 & 5136 & 4.4\% & 3.9\% \\
 & Sonnet-5 & 133 & 1433 & 1.0\% & 1.9\% \\
 & GPT-5.5 & 211 & 5369 & 7.2\% & 3.2\% \\
 & Qwen3.5-9B & 136 & 2814 & 0.7\% & 0.2\% \\
\midrule
Airline & GPT-5.4-mini & 21 & 207 & 0.5\% & 1.2\% \\
(policy) & Sonnet-5 & 42 & 242 & 2.0\% & 2.9\% \\
 & Qwen3.5-9B & 35 & 191 & 0.0\% & 0.3\% \\
\midrule
Airline & GPT-5.4-mini & 26 & 5043 & 14.4\% & 9.4\% \\
(no policy) & Sonnet-5 & 46 & 5061 & 13.0\% & 12.3\% \\
 & Qwen3.5-9B & 41 & 2448 & 4.2\% & 1.5\% \\
\midrule
Retail & GPT-5.4-mini & 17 & 63 & 0.5\% & 1.0\% \\
(policy) & Sonnet-5 & 21 & 61 & 0.4\% & 1.1\% \\
 & Qwen3.5-9B & 10 & 50 & 0.7\% & 0.2\% \\
\midrule
Retail & GPT-5.4-mini & 17 & 65 & 0.0\% & 0.6\% \\
(no policy) & Sonnet-5 & 21 & 60 & 0.0\% & 0.7\% \\
 & Qwen3.5-9B & 10 & 51 & 0.0\% & 0.1\% \\
\bottomrule
\end{tabular}
\caption{Checker cost per tool call over the no-constraint runs (AppWorld: 417 tasks; $\tau^2$-bench: 20 airline and 40 retail tasks).
p95 is the 95th-percentile check time.
A timeout is a check where Z3 returns unknown at the 5\,s limit.
Share of episode time is total check time divided by episode wall-clock time, averaged over tasks.}
\label{tab:solver-cost}
\end{table}

\paragraph{Cost.}
The cost of checking is negligible relative to the LLM agent (Table~\ref{tab:solver-cost}). The median check takes 10--350\,ms, and the checker accounts for 0.1--4\% of episode wall-clock time on every catalog except airline without policy. On that catalog, Z3 times out on up to 14\% of checks, which costs up to 12\% of episode time. The live checks during the AppWorld warning and repair runs agree: a median of 92--248\,ms per call (p95 0.8--5.6\,s), or 0.3--3.2\% of episode time. Check time grows with the length of the call history rather than with catalog size. On AppWorld, the median check takes 58\,ms with fewer than 10 prior calls and 2.6\,s beyond 80.

\end{document}

%% file: intro.tex
\section{Introduction}

Large language model (LLM)-based agents are increasingly used to automate real-world tasks, from customer service to complex workspace duties \citep{trivedi_appworld_2024, gao_tool_2025, fang_toolatlas_2026}. A critical component of these agents is \textit{tools}, or external functions that allow agents to read information from the environment and update the environment state \citep{schick_toolformer_2023}. As agents interact with increasingly complex environments, how can we ensure that their tool calls are valid?


Previous work primarily focuses on formalizing predefined safety policies into symbolic constraints and enforcing them on all tool calls by blocking violations for better policy adherence \citep{kamath_enforcing_2025, zwerdling_towards_2025}. However, complex tool-calling environments impose many rules that are not explicitly codified as policies. These include dependencies between tool calls, argument provenance requirements, and restrictions on the current environment state \citep{trivedi_appworld_2024, shi_retrieval_2025}. Although some of these conditions can be inferred from API documentation, others become apparent only through successful or failed interactions with the environment. 


\begin{figure}[!tp]
    \centering
    \includegraphics[width=0.8\linewidth]{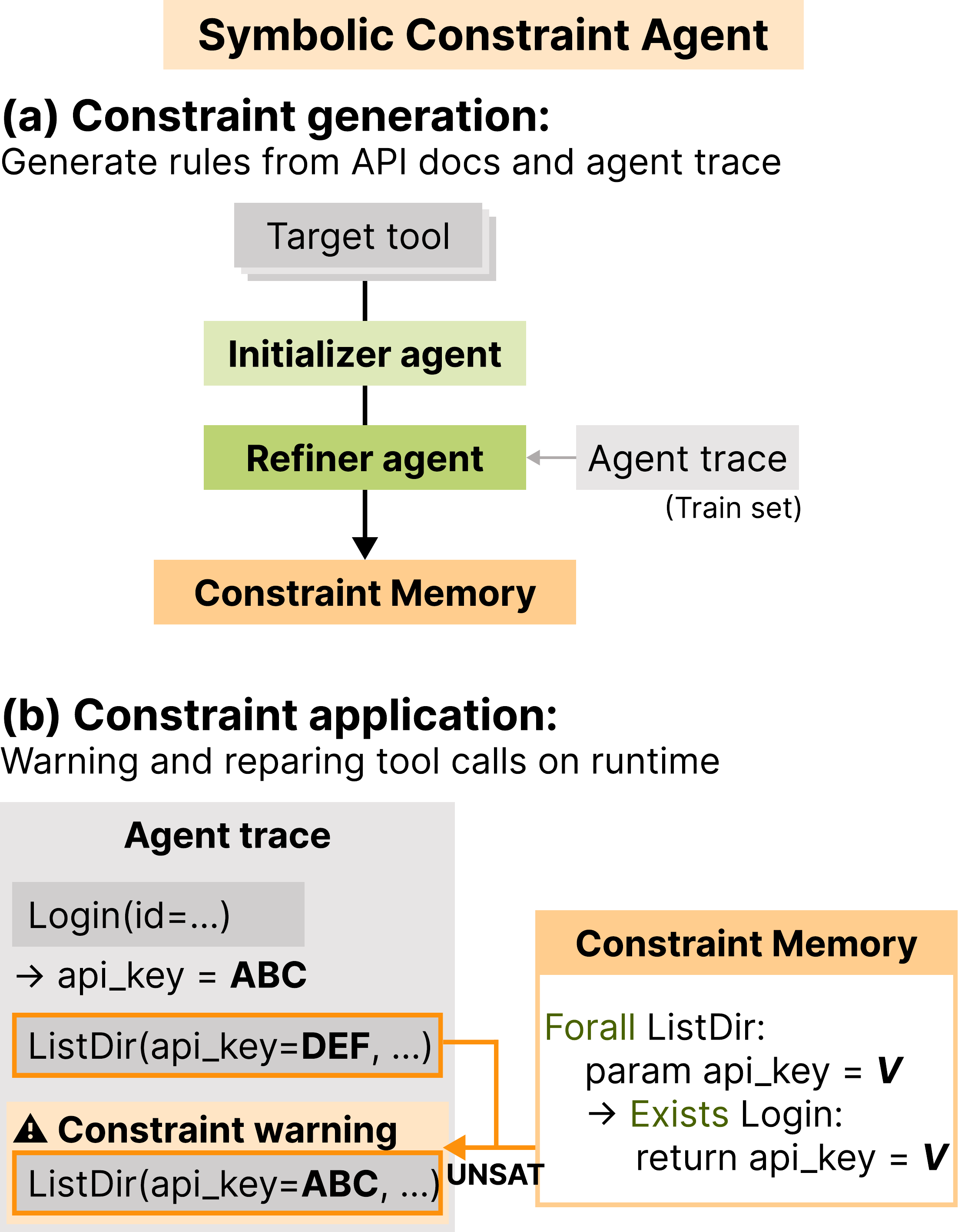}
    \caption{Overview of SCA. Constraint generation (top) runs once per environment: an initializer agent mines candidate rules from API documentation and a refiner agent corrects them against sample agent traces. At the constraint application stage (bottom), each proposed call is checked against the constraint memory using an SMT solver. The rule shown requires \textsc{ListDir}'s \texttt{api\_key} to originate from an earlier
    \textsc{Login}; the call passing \texttt{DEF} is unsatisfiable and can be repaired to \texttt{ABC}.}
    \label{fig:main}
\end{figure}

This raises two questions: How can we generate executable tool-call constraints without manually specified policies, and can these constraints improve agents' ability to complete tasks rather than merely enforce compliance? Addressing these questions requires a coherent system that mines constraints from diverse sources like policy and tool documentation, refines inaccurate rules using observed execution behavior, and translates detected violations into runtime interventions.

We propose Symbolic Constraint Agents (SCA), which generate tool dependency constraints as first-order logic rules and verify agents' tool calls with SMT (Satisfiability Modulo Theories) solvers~\citep{demoura_z3_2008} when each call is executed in complex tool environments. First, an initializer agent mines candidate rules from the API documentation. Since relying solely on API documents yields constraints that do not reflect the real environment, a second refiner agent removes overly strict rules and adds missing rules using sample agent traces. At runtime, the agent receives information about violated constraints each turn, guiding the agent to retry. Furthermore, for constraint violations with a unique correction (e.g., passwords), SCA can instead repair the broken argument using MaxSMT model checking.

\begin{figure*}[!tp]
    \centering
    \includegraphics[width=\linewidth]{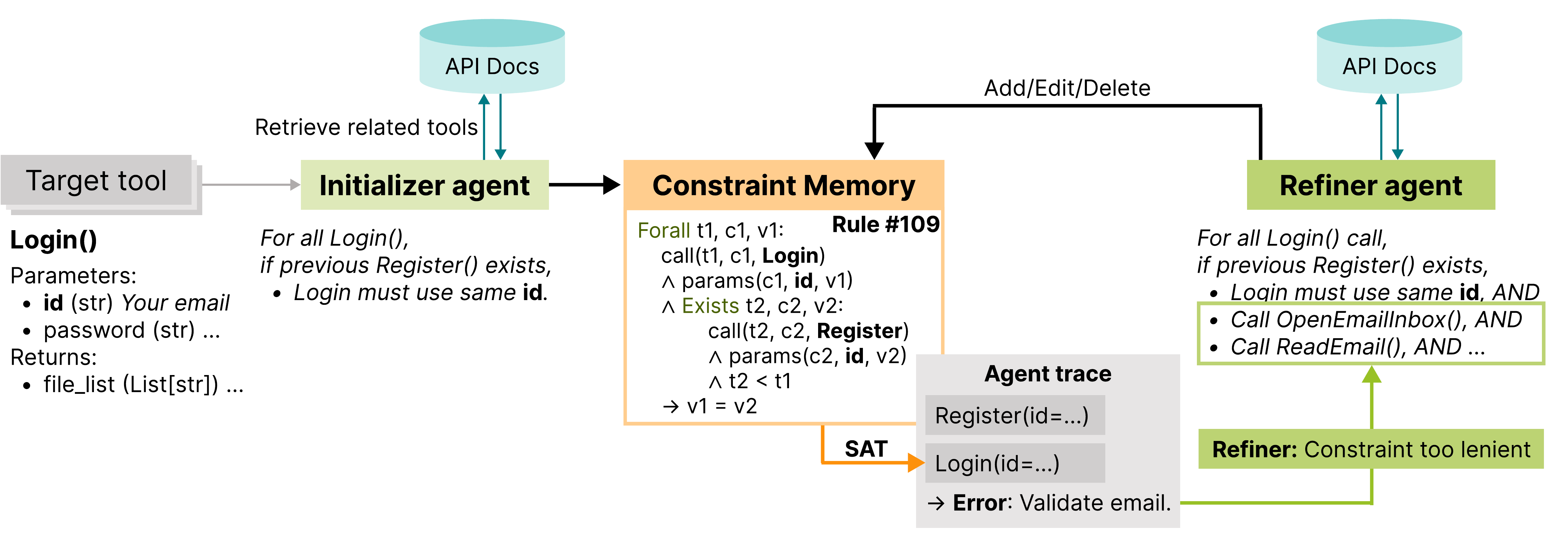}
    \caption{Constraint generation. \textbf{Left:} Given the target tool's specification, the initializer agent generates constraints by reasoning about the given tool and searching other potentially relevant tools' definitions. \textbf{Right:} After the initializer agent, the refiner agent updates the catalog based on how the current constraints apply to agent traces.}
    \label{fig:constraint-generation}
\end{figure*}

%

We make three contributions. First, we introduce an environment-grounded constraint mining framework that mines executable tool-call rules from API documentation and iteratively refines them against execution traces, even without requiring manually specified task-success criteria. Second, we develop solver-based interventions that translate detected violations into natural-language feedback, or automatic repairs for uniquely determined argument errors. Third, we demonstrate that symbolic constraints can improve end-to-end task completion through experiments on AppWorld~\citep{trivedi_appworld_2024} and $\tau^2$-bench~\citep{barres_tau2-bench_2025}, particularly when tool-call failures are costly or hard to diagnose, while reducing failed calls and inference costs.

%% file: related-work.tex
\section{Related Work}

\paragraph{Tool-calling environments.} Tool-calling environments evaluate agents' ability to manage persistent state, cross-application dependencies, and long action sequences. Benchmarks like $\tau^2$-Bench \citep{barres_tau2-bench_2025} focus on evaluating policy adherence and user intent clarification via conversation, while benchmarks like AppWorld and TheMCPCompany focus on tool planning and execution for completing complex real-world tasks \citep{trivedi_appworld_2024, esfandiarpoor_themcpcompany_2025, wang_mcp-bench_2025}. We evaluate on both kinds: AppWorld, whose large tool catalog, cross-application tasks, and executable evaluator make tool dependencies both necessary and measurable, and $\tau^2$-bench, where the dependencies come from a written policy and many violations go unreported by the environment.



\paragraph{Symbolic constraints for tool-calling agents.} Agent-C \citep{kamath_enforcing_2025} and ToolGuard \citep{zwerdling_towards_2025} translate safety policies into first-order constraints, symbolically check whether each tool call violates the constraints, and apply constrained generation or resampling to produce safe tool calls. Recent systems vary in what a human must supply: AgentSpec \citep{wang2025agentspec} prompts an LLM with developer-written seed rules, AgentLTL \citep{elkoussy2026agentltl} instantiates constraints from a gold trace per task, and AutoSpec \citep{ma2026autospec} refines an expert-authored guardrail against safety-labelled traces.
Instead, ours dynamically learns executable constraints from the environment's response, minimizing human intervention and generalizing beyond preexisting policies.

These works do not scale to environments with hundreds of tools and no compliance criterion
\citep{trivedi_appworld_2024, esfandiarpoor_themcpcompany_2025}, because each still requires a constraint artifact authored by hand for the target environment. 
None evaluates whether constraints improve task completion either; for instance, Agent-C reports gains only when a policy violation counts as a failure.

\paragraph{Memory for complex tool-calling environments.}
With many tools, agents typically retrieve what they need from API documentation on the fly \citep{qin_toolllm_2024}, so nothing learned in one episode carries to the next.
This motivates treating tool knowledge as a reusable \emph{memory} rather than as a per-query retrieval problem.
Initial work \citep{gao_tool_2025, lee_-n-out_2025} represented tool dependencies as directed graphs and used them as memory to augment relevant tool retrieval. However, these graphs are obtained by expert annotation or a fine-tuned classifier, and remain static between episodes. This prevents the agent from actively updating and modifying the knowledge about the environment. ToolAtlas \citep{fang_toolatlas_2026} pushes the abstraction further by maintaining a persistent, provider-side tool memory graph through exploration and reuses it across agents and tasks. Symbolic Constraint Agent maintains a dynamic and executable memory based on \textit{formal logic}: tool-calling rules can be updated using the environment's signals and can directly check whether a tool-call attempt is valid.

%% file: symbolic-agents.tex
\section{Symbolic Constraint Agents}

\subsection{Expressing tool calls in formal logic}
\label{subsec:formulation}

Tool calls are sequential activities, which we model using first-order logic. Specifically, we define the sequence of tool calls using three predicates (uninterpreted functions).

\begin{center}
\begin{tabularx}{\linewidth}{@{}l@{\quad}X@{}}
  $\operatorname{call}(t,c,f)$ & tool $f$ is invoked at time $t$, with unique call ID $c$,\\
  $\operatorname{params}(c,k,v)$ & argument $k$ of $c$ has value $v$,\\
  $\operatorname{returns}(c,k,v)$ & return field $k$ of $c$ has value $v$.
\end{tabularx}
\end{center}
Values $v$ are of primitive types including strings, integers, reals, Booleans, and null. Complex data structures, including maps and lists, are unrolled so each primitive value is attached to the closest key. Tool and field names are finite symbolic domains.

We introduce closed-world assumption statements to prevent the SMT solver from inventing unasserted facts about the tool-calling history. First, we assume that for each timestep $t$, there exists a unique $\operatorname{call}(t,c,f)$. Then, we assume that for all past tool calls, parameters, and return values are final; that is, for every call ID $c$, no $\operatorname{params}$ or $\operatorname{returns}$ predicates with $c$ exist other than the conjunction of existing predicates. ~\autoref{app:z3} gives the full sorts, predicates, and axioms. Rules over the call history or the call's outcome account for 98\% of flagged calls. 

Under this formulation, constraints are formulated as universally quantified rules. For example, the constraint ``If \textsc{Register} tool was called, a subsequent \textsc{Login} must use the same id as used for \textsc{Register}" can be expressed as
\begin{align*}
&\forall t_1,c_1,v_1,t_2,c_2,v_2.\\
&\quad \operatorname{call}(t_1,c_1,\textsc{Login}) \land \operatorname{params}(c_1,\textsc{Id},v_1)\\
&\quad \land \operatorname{call}(t_2,c_2,\textsc{Register}) \land \operatorname{params}(c_2,\textsc{Id},v_2)\\
&\quad \land t_2<t_1 \;\rightarrow\; v_1 = v_2.
\end{align*}
\normalsize

If the constraint is unsatisfiable (UNSAT) under the current tool call facts, it indicates a violation. For instance, the example above is unsatisfiable if the \textsc{Login} tool uses a different \textsc{Id} parameter than what the previous \textsc{Register} call returns.
A naive cumulative check can repeatedly report a contradiction caused by an earlier call. We therefore compute an unsatisfiable core and intervene only when at least one rule in the core is instantiated by the \emph{current} call. Since solvers do not guarantee a minimal core, this attribution is approximate rather than exact. However, we observe that multiple constraints being violated together is rare.

\subsection{Constraint generation}
\label{subsec:constgen}

The constraint generation process mines constraints across different tools, including parameter value dependencies and environment state dependencies.

\paragraph{Initialization from documentation.} The \textit{initializer agent} generates constraints for one target API at a time. Since the full tool collection cannot fit in a model's context window, retrieval supplies the target definition together with related tools selected by name and description. From this context, the model identifies candidate preconditions, such as value provenance (e.g., an identifier must originate from an earlier lookup), temporal ordering (e.g., authentication must precede a protected operation), type and domain restrictions (e.g., a quantity must be positive or a status must come from a fixed set of values), and observable-state preconditions (e.g., a requested amount must not exceed the available inventory). For each candidate, it emits both a natural-language description and a corresponding Z3 expression. For example, in Figure~\ref{fig:constraint-generation}, the target tool is \textsc{Login}. Retrieval brings in the \textsc{Register} definition, and the initializer produces a rule
requiring a \textsc{Login} to reuse the \texttt{id} from an earlier \textsc{Register}, along with the description ``\textsc{Login} must use same id as used for \textsc{Register}.''

Constraint generation proceeds incrementally across target APIs rather than processing each API independently: the model can attach a tool to an existing parameter-level rule, create a new rule, edit a rule shared across related tools, or remove one found to be invalid, retrieving additional API definitions whenever a dependency is unclear. Before entering the catalog, every candidate expression is parsed and type-checked.
(Prompts in ~\autoref{app:prompts}).

\begin{figure}[!tp]
    \centering
    \includegraphics[width=0.9\linewidth]{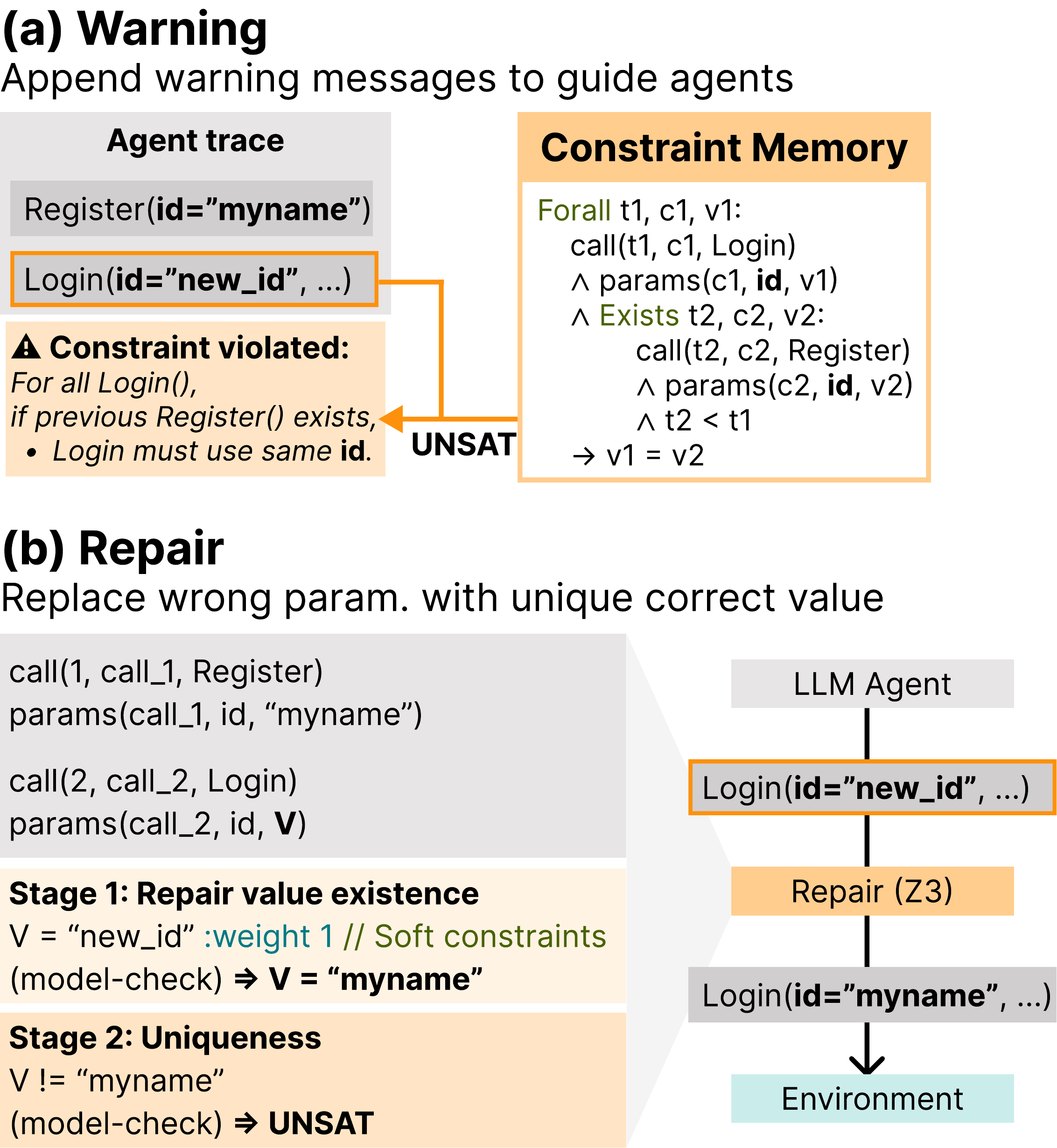}
\caption{Constraint application. \textbf{(a)} In warning mode, the call is left unmodified, and the violated constraint is appended to the agent's context in natural language. Here, a
\textsc{Login} that uses a different \texttt{id} from the preceding \textsc{Register}. \textbf{(b)} In repair mode, SCA treats each argument as a soft constraint and finds the assignment with minimal change (Stage 1), then asserts $v \neq v^\star$ and re-checks to confirm the replacement is unique (Stage 2). The repaired call is then passed to the environment.} \label{fig:constraint-application}
\end{figure}

\paragraph{Refinement from traces.} Documentation-derived constraints generated by the initializer agent are likely imperfect. They might be too strict by rejecting a successful tool call, or too lenient when no rule accounts for an environment-rejected call. 
This discrepancy between the generated constraints and the real environment can be easily found by applying the constraints to actual agent execution traces. The \textit{Refiner agent} updates the set of constraints by sampling a trajectory from the training set. It receives the full agent trajectory with tool calls and return values, along with the satisfiability results and UNSAT core for each tool call obtained from the current set of constraints. It then reasons about whether each step's verdict (UNSAT or SAT) is valid using the provided information. The refiner agent then proposes edits to the constraint memory by adding rules that capture failures that slipped through and modifying rules that are overly restrictive on correct tool calls. The results of applying the edited constraints to the current trace are returned to the agent, and the refiner agent is allowed to propose edits multiple times until it decides to terminate or exceeds the maximum number of tool calls\footnote{Turn limits were set to 20 in practice. However, none of the refiner agent's attempts exceeded this limit.}.

To ensure fair comparison with baselines, we do not provide the oracle evaluation results of the input trace (whether the trace fulfilled the user's intent or not), and leave the judgment to the refiner agent. This design alleviates the burden of authoring and validating evaluation criteria for the training tasks, and allows integration with data augmentation methods for better tool coverage.

\subsection{Constraint application to agents}
\label{subsec:application}

After refinement, we apply the symbolic constraints to the agent runtime to improve performance.

\paragraph{Warning.} In warning mode, SCA checks whether any constraint was violated \textit{after} the environment returns the tool execution result, allowing constraints to process both parameter-side errors and return value-based recommendations. It leaves the call unmodified and appends the violated natural-language constraints to the agent context. The model can then revise its plan or arguments. 
Warnings preserve agent autonomy and are appropriate when corrections or alternative plans may be valid; ~\autoref{fig:warning-example} shows a recorded instance.

\paragraph{Repair.} Some invalid arguments have a unique correct value already present in the trajectory, especially service-specific access tokens and entity identifiers. In repair mode, SCA first detects an unsatisfiable constraint \textit{before} the tool call is made, treats each argument leaf as a candidate change, and uses MaxSMT~\citep{bjorner2015nuz} to find a minimum-change assignment that satisfies the previously violated constraints. ~\autoref{fig:repair-example} shows a recorded repair of a mistyped access token.

Repairs are applied conservatively only when the replacement is unique, to prevent malformed repairs that do not preserve user intent. After obtaining a candidate value $v^\star$ via MaxSMT, SCA asserts $v\neq v^\star$ and runs the solver again. If $v\neq v^\star$ is unsatisfiable, $v^\star$ is the only possible replacement and is then passed to the environment. If more than one value remains possible, SCA falls back to a warning rather than silently choosing among valid alternatives. If the constraints' verdict was UNSAT but without a unique repair, it behaves as in warning mode, appending violated constraints to the agent's context.

%% file: eval.tex
\section{Experimental Setup}
\label{sec:setup}
 \paragraph{Benchmark.} We evaluate SCA on two tool-calling benchmarks with different error structures.
 
 \textbf{AppWorld} \citep{trivedi_appworld_2024}: we use the \texttt{test-challenge} set (417 instructions). The AppWorld environment exposes 457 APIs across nine diverse stateful applications, including email, music, notes, and contacts. Using these tools, the agent should complete users' tasks such as accepting Venmo transactions and sending emails to friends. The default environment provides an informative error message whenever the tool call fails due to incorrect parameter values, authentication error, etc. (\emph{Standard}) However, an agent can often recover from a rejected call by reading the error message and retrying, so a prevented error may only save cost.
We therefore introduce two challenging variants of AppWorld that make tool errors increasingly expensive: 
\emph{40-call budget}, where the agent may make at most 40 tool calls instead of the Standard setting's 300, so every call spent on an error is lost to the task;
and \emph{silent errors}, where the environment rejects erroneous calls with an empty response, which are harder for agents to recover from failures \citep{zhang_robust_2026}. For the silent errors setting, the constraint generation/refinement stage also does not access the full error message.

\textbf{$\tau^2$-bench} \citep{barres_tau2-bench_2025}: we use the airline and retail domains with 20 and 40 held-out tasks, respectively. The agent serves an LLM-simulated customer through booking and order-management tools while following a written company policy, and the official evaluator checks the final database state and the required actions. 
Unlike AppWorld, the environment does not reject malformed tool calls, including ones that violate the policy (e.g., cancelling a reservation that the policy does not allow to be cancelled), making it difficult for agents to detect their own errors.

For tool navigation, the AppWorld environment provides a default BM25-based search tool as a separate app. Initializer and Refiner agents in SCA's constraint generation stage utilize this endpoint to search for relevant tools when generating/editing constraints. $\tau^2$-bench domains have few tools (14 in airline, 16 in retail), so the initializer and refiner receive all tool definitions along with the full policy in its context.

\paragraph{Constraint refinement data.} Refinement uses the training set of AppWorld and $\tau^2$-bench, containing 90 and 104 prompts, respectively. We use plain tool-calling LLM agents' tool-calling histories to identify discrepancies between the current constraint catalog and the actual environment, then freeze the refined catalog before applying it to the test set. 
We use GPT-5.5 for both initializer and refiner agents. The refined catalogs contain 295 constraints for AppWorld, 51 for airline, and 63 for retail (\autoref{app:catalogs}). We do not use the oracle grader's task-success evaluations to ensure SCA's generalizability when such graders are unavailable.

\paragraph{Models and conditions.}
We evaluate GPT-5.5 and GPT-5.4-mini \citep{openai_gpt54mini_2026}, Sonnet 5 \citep{anthropic_sonnet5_2026}, Qwen3.5-9B and Qwen3.5-27B \citep{qwen_qwen35_2026} as agent models; Qwen3.5-27B is evaluated in Appendix~\ref{app:qwen27b}.
For each model, we compare standard tool-calling mode, SCA with warnings, and SCA with automatic repair. Runs share the task evaluator, agent harness, and tool infrastructure (\autoref{app:implementation}); only the symbolic intervention mode changes.

%% file: results.tex
\section{Results}

\begin{table}[t]
  \centering
    \footnotesize
  \begin{tabular}{lccc}
    \toprule
    & Precision (\%) & Recall (\%)   & F1 (\%)       \\
    \midrule
    \multicolumn{4}{l}{\textit{GPT-5.5 traces}}                          \\
    \quad Initialize    & 61.5           & 26.4          & 37.0          \\
    \quad + Refine      & \textbf{90.4}  & \textbf{43.0} & \textbf{58.3} \\
    \midrule
    \multicolumn{4}{l}{\textit{Qwen3.5-27B traces}}                      \\
    \quad Initialize    & 58.8           & 29.6          & 39.3          \\
    \quad + Refine      & \textbf{83.9}  & \textbf{49.1} & \textbf{61.9} \\
    \bottomrule
  \end{tabular}
  \caption{Detection of environment-rejected calls. Refinement improves precision and recall on trajectories from both model families.}
  \label{tab:error-detection}
\end{table}

\subsection{Quality of Generated Constraints}
\label{subsec:error-detection}

\paragraph{Tool-call error detection.} First, we evaluate whether SCA can detect faulty calls that the environment will reject. The AppWorld environment returns an error message whenever a tool call is invalid for any reason, e.g., authentication errors, an empty cart when submitting an order, etc. SCA's proposed constraints can therefore be evaluated via \textit{binary classification}: the ground truth is whether a tool call returns an error, and the prediction is whether a constraint is violated at the given step.\footnote{We run this evaluation on AppWorld only, as the $\tau^2$-bench environment does not reject most invalid calls and thus the environment signal is not trustworthy.}



\begin{table*}[t]
  \centering
  \small
  \setlength{\tabcolsep}{3pt}
  \newcommand{\ch}[3]{#1\,$\to$\,#2 {\scriptsize(#3)}}
  \begin{tabular}{llccc|cc}
    \toprule
    & & \multicolumn{3}{c|}{Task success (\%)} & \multicolumn{2}{c}{Failed calls per task} \\
    \cmidrule(lr){3-5} \cmidrule(lr){6-7}
    Benchmark (Setting) & Agent model & No constraints & + Warning & + Repair & + Warning & + Repair \\
    \midrule
    AppWorld (Standard)       & Qwen3.5-9B   & 18.1 $\pm$ 2.2 & 17.6 $\pm$ 1.2           & \textbf{20.4} $\pm$ 2.0  & \ch{4.6}{1.1}{$\mathbf{-76}$} & \ch{4.6}{0.6}{$\mathbf{-87}$} \\
    AppWorld (Standard)       & GPT-5.4-mini & 73.5 $\pm$ 1.5 & 71.2 $\pm$ 2.0           & \textbf{75.1} $\pm$ 3.7  & \ch{6.6}{5.1}{$-$24}   & \ch{6.6}{2.7}{$\mathbf{-59}$}  \\
    AppWorld (Standard)       & Sonnet-5     & 77.3 $\pm$ 2.9 & 75.8 $\pm$ 1.7           & \textbf{78.1} $\pm$ 0.5  & \ch{1.0}{0.9}{$-$5}    & \ch{1.0}{0.8}{$-$14}           \\
    AppWorld (Standard)       & GPT-5.5      & 90.8 $\pm$ 0.8 & \textbf{91.1} $\pm$ 1.0  & \textbf{91.4} $\pm$ 1.0  & \ch{0.95}{0.97}{+3}    & \ch{0.95}{0.74}{$-$22}         \\
    \midrule
    AppWorld (40-call budget) & Qwen3.5-9B   & 12.5 $\pm$ 0.0 & 11.1 $\pm$ 0.6           & \textbf{12.8} $\pm$ 0.1  & \ch{1.1}{0.61}{$-$43}  & \ch{1.1}{0.56}{$-$48}          \\
    AppWorld (40-call budget) & GPT-5.4-mini & 45.1 $\pm$ 0.7 & 44.5 $\pm$ 2.5           & \textbf{50.5} $\pm$ 0.5* & \ch{1.8}{1.7}{$-$8}    & \ch{1.8}{0.7}{$\mathbf{-59}$}  \\
    AppWorld (40-call budget) & Sonnet-5     & 60.9 $\pm$ 1.7 & \textbf{64.0} $\pm$ 0.3* & \textbf{61.2} $\pm$ 3.7  & \ch{0.4}{0.4}{$-$7}    & \ch{0.4}{0.4}{$-$5}            \\
    AppWorld (40-call budget) & GPT-5.5      & 65.3 $\pm$ 1.2 & 65.0 $\pm$ 0.3           & 65.1 $\pm$ 1.5           & \ch{0.57}{0.56}{$-$2}  & \ch{0.57}{0.45}{$-$21}         \\
    \midrule
    AppWorld (Silent errors)  & Qwen3.5-9B   & 19.3 $\pm$ 0.2 & \textbf{20.5} $\pm$ 0.8  & \textbf{20.3} $\pm$ 0.5  & \ch{9.7}{5.6}{$-$43}   & \ch{9.7}{2.8}{$\mathbf{-71}$}  \\
    AppWorld (Silent errors)  & GPT-5.4-mini & 59.5 $\pm$ 1.7 & \textbf{65.2} $\pm$ 1.7* & \textbf{69.3} $\pm$ 1.0* & \ch{7.4}{4.5}{$-$40}   & \ch{7.4}{2.5}{$\mathbf{-67}$}  \\
    AppWorld (Silent errors)  & Sonnet-5     & 77.8 $\pm$ 0.8 & 75.8 $\pm$ 1.7           & 76.1 $\pm$ 1.2           & \ch{1.3}{1.4}{+3}      & \ch{1.3}{1.2}{$-$6}            \\
    AppWorld (Silent errors)  & GPT-5.5      & 90.4 $\pm$ 0.3 & 89.6 $\pm$ 0.5           & 90.4 $\pm$ 0.3           & \ch{0.97}{0.78}{$-$20} & \ch{0.97}{0.82}{$-$16}         \\
    \midrule
    $\tau^2$-bench     & Qwen3.5-9B   & 23.9 $\pm$ 4.2 & \textbf{28.3} $\pm$ 3.3 & \textbf{29.4} $\pm$ 2.5 & \ch{0.40}{0.49}{+23}   & \ch{0.40}{0.32}{$-$21} \\
    $\tau^2$-bench     & GPT-5.4-mini & 67.2 $\pm$ 2.5 & \textbf{68.9} $\pm$ 4.2 & \textbf{72.8} $\pm$ 1.0 & \ch{0.14}{0.10}{$-$28} & \ch{0.14}{0.09}{$-$36} \\
    $\tau^2$-bench     & Sonnet-5     & 83.3 $\pm$ 2.9 & 80.0 $\pm$ 1.7          & 81.1 $\pm$ 4.8          & \ch{0.12}{0.11}{$-$9}  & \ch{0.12}{0.11}{$-$13} \\
    \bottomrule
  \end{tabular}
  \caption{Task success and failed tool calls.
  AppWorld: 417 test tasks; $\tau^2$-bench: 60 test tasks (airline and retail); mean $\pm$ standard deviation over three runs.
  Success: bold marks an improvement over no constraints, and * marks a statistically significant difference (paired task bootstrap, $p < 0.05$).
  Failed calls: calls per task rejected by the environment, without constraints $\to$ with the given mode (change in \%); bold marks reductions of 50\% or more.}
  \label{tab:main}
\end{table*}

\begin{table}[!t]
  \centering
  \footnotesize
  \setlength{\tabcolsep}{1pt}
  \newcommand{\cp}[3]{#1$\to$#2\,{\scriptsize(#3)}}
  \newcommand{\cm}[3]{\cellcolor{blue!12}#1$\to$#2\,{\scriptsize($-$#3)}}
  \newcommand{\chh}[3]{\cellcolor{blue!30}#1$\to$#2\,{\scriptsize($-$#3)}}
  \begin{tabular}{l ccc}
    \toprule
    Per task     & Standard            & 40-call             & Silent               \\
    \midrule
\multicolumn{4}{l}{\textit{Qwen3.5-9B}} \\
Tool calls   & \cp{32}{29}{$-$9}     & \cp{22}{22}{$-$2}     & \cm{46}{38}{17}        \\
LLM calls    & \cp{45}{43}{$-$3}      & \cp{30}{30}{0}        & \cp{68}{62}{$-$10}     \\
Prompt (k)   & \cp{954}{869}{$-$9}    & \cp{514}{519}{+1}     & \cp{1428}{1291}{$-$10} \\
Compl.\ (k)  & \cm{23.8}{21.4}{10} & \cp{11.9}{13.7}{+15}  & \cm{16.1}{13.8}{14}    \\
\midrule
    \multicolumn{4}{l}{\textit{GPT-5.4-mini}} \\
    Tool calls   & \cm{55}{48}{14}     & \cp{34}{33}{$-$1}   & \cm{51}{45}{13}      \\
    LLM calls    & \cm{26}{21}{20}     & \cp{14}{14}{0}      & \cm{22}{19}{13}      \\
    Prompt (k)   & \cm{840}{679}{19}   & \cp{358}{365}{+2}   & \cm{696}{606}{13}    \\
    Compl.\ (k)  & \cm{25.8}{18.8}{27} & \cp{8.8}{8.5}{$-$3} & \cm{17.7}{14.5}{18}  \\
    \midrule
    \multicolumn{4}{l}{\textit{Sonnet-5}} \\
    Tool calls   & \cp{37}{36}{$-$2}   & \cp{30}{30}{0}      & \cp{38}{38}{$-$1}    \\
    LLM calls    & \cp{23}{23}{0}      & \cp{20}{20}{0}      & \cp{24}{24}{$-$1}    \\
    Prompt (k)   & \cp{822}{817}{$-$1} & \cp{640}{643}{0}    & \cp{914}{898}{$-$2}  \\
    Compl.\ (k)  & \cp{7.7}{7.4}{$-$4} & \cp{5.8}{5.8}{0}    & \cp{7.9}{7.9}{0}     \\
    \midrule
    \multicolumn{4}{l}{\textit{GPT-5.5}} \\
    Tool calls   & \cp{43}{42}{$-$1}   & \cp{33}{33}{0}      & \cp{42}{42}{0}       \\
    LLM calls    & \cp{16}{16}{0}      & \cp{12}{11}{$-$1}   & \cp{15}{15}{$-$1}    \\
    Prompt (k)   & \cp{546}{544}{0}    & \cp{315}{310}{$-$1} & \cp{488}{481}{$-$1}  \\
    Compl.\ (k)  & \cp{7.4}{7.3}{$-$1} & \cp{4.8}{4.8}{0}    & \cp{7.1}{7.1}{0}     \\
   \bottomrule
  \end{tabular}
  \caption{Cost per AppWorld task without constraints $\to$ repair (change in \%).
  \emph{Prompt}/\emph{Compl.}: prompt and completion tokens.
  Shading: savings more than 10\%.}
  \label{tab:cost}
\end{table}

Table~\ref{tab:error-detection} reports precision, recall, and F1 on trajectories generated by GPT-5.5 and Qwen3.5-27B, capturing both a strong proprietary model and a weaker open-weight model.
The results show that using tool definitions alone (Initialize) produces noisy constraints. Refinement with a small number of traces raises precision by 28.9 points on GPT-5.5 trajectories and 25.1 points on Qwen3.5-27B trajectories in the test set, while recall increases by 16.6 and 19.5 points, respectively. This shows the necessity and effectiveness of the refinement stage.

Maximum recall remains lower than precision, primarily because some errors depend on environmental conditions that cannot be inferred from tool documentation or agent trace alone. For instance, in AppWorld, whether a file actually exists in the environment cannot be determined in advance. Consequently, the constraints cannot (and should not be expected to) detect tool calls that reference nonexistent files. Such calls are rejected by the environment and counted as false negatives, limiting achievable recall.

We further audit the constraint catalogs mined from AppWorld and $\tau^2$-bench using LLMs as a judge. The results show that around 92\% of the rules are \textit{valid}, as they define constraints that accurately reflect tool description/environment behaviors, and \textit{faithful} to their natural language descriptions. Further details, including manual inspections to measure calibration between human and LLM judges, can be found in \autoref{app:rule-audit}.

\subsection{Task Success and Efficiency}

We compare three agents that differ only in how the constraints are used: \emph{no constraints} (the vanilla tool-calling agent), \emph{warning} (after each call, violated constraints are reported to the agent in the tool response), and \emph{repair} (before each call, a violating argument is replaced by the value that satisfies the constraints, when a unique such value exists in the call history).
All constraint runs use the same refined catalog per benchmark, built once with GPT-5.5 and frozen.
We evaluate three proprietary models (GPT-5.5, GPT-5.4-mini, Sonnet-5) and the open-weight Qwen3.5-9B. We present single-run results for Qwen3.5-27B in Appendix~\ref{app:qwen27b}, which follows similar performance improvement trends.

%
Table~\ref{tab:main} reports mean task success for every (model, setting, mode) combination, along with the standard deviation over three independent runs. Table~\ref{tab:cost} reports the per-task cost of the same runs.

\mypara{Constraints improve task success, most when steps are limited}
Constraints improve task success for GPT-5.4-mini, Sonnet-5 and Qwen3.5-9B in the standard and budget-limited AppWorld settings with at least one of the two modes (Table~\ref{tab:main}); for GPT-5.5, which already solves most tasks, success rates remain stable.
In the standard setting with up to 300 calls, agents can usually recover from a rejected call by retrying. This gives relatively small gains: repair raises success rate from 73.5\% to 75.1\% for GPT-5.4-mini and from 77.3\% to 78.1\% for Sonnet-5.
Warnings can even cost here: they lower GPT-5.4-mini and Sonnet-5 by 1.5--2.3 points, since a false alarm makes an agent retry a valid call.
Therefore, in the unrestricted setting, preventing an error mainly saves cost (Table~\ref{tab:cost}).
%
Under a tight 40-call budget, calls cannot be wasted recovering from errors, so preventing errors leads to higher success.
Constraints improve the performance for both models: repair improves GPT-5.4-mini from 45.1\% to 50.5\% (+5.4 points), and warning improves Sonnet-5 from 60.9\% to 64.0\% (+3.1 points).
GPT-5.5 does not gain under the budget, although repair again removes 21\% of its failed calls; it makes few errors to begin with, so preventing them frees little budget.
The gains concentrate in mid-strength models, which make enough errors for constraints to catch and have budget left to use the corrections.

\mypara{Constraints lead to fewer failed calls and lower cost}
Because nearly every call that violates a constraint is one the environment would reject (Table~\ref{tab:error-detection}), repair effectively prevents failures before they happen: it removes 59--67\% of failed calls for GPT-5.4-mini and 48--87\% for Qwen3.5-9B across different settings.
Each prevented failure also saves the turns the agent would have spent recovering, so GPT-5.4-mini needs 20\% fewer LLM calls and 27\% fewer completion tokens per task in the standard setting, consistently across all three runs (Table~\ref{tab:cost}).
The checker's cost is mostly negligible; Z3 checking adds a median of 0.1--0.4\,s per call (Details in Appendix~\ref{sec:solver}).

\mypara{When errors are silent, constraints can restore the missing signal}
When the environment returns blank error messages, GPT-5.4-mini costs 14 points (73.5\% $\rightarrow$ 59.5\%) in AppWorld end-to-end performance, showing its low ability to recover from error.
However, the constraints can still flag the faulty call with high precision, even though the constraint generation/refinement stage receives blank error messages.
In GPT-5.4, warning mode recovers 5.7 points and repair recovers 9.8 points ($p < 0.05$ for both), which takes account of about 70\% of the loss compared to the Standard setting. The number of failed calls that go unnoticed also falls from 7.4 per task to 4.5 with warnings and 2.5 with repair, signaling strong improvement.
%

$\tau^2$-bench also provides uninformative error messages, making it comparable to the Silent errors setting of Appworld.
With constraints that explicitize domain policy and tool contracts, repair raises success by 5.6 points for GPT-5.4-mini and 5.5 points for Qwen3.5-9B, and warnings raise Qwen3.5-9B by 4.4 points.

\begin{figure}[]
\small
\begin{transcript}{Warning:}
\trow{User:}{\textit{"Place an order for everything in my Amazon cart and wishlist, shipped to my home address."}}
\trow{}{...}
\trow{Agent:}{\texttt{place\_order(payment\_card\_id=64, address\_id=29,\allowbreak{} access\_token="0ac87b...")}}
\trow{Env:}{Invalid promo code.}
\scarow{Warning:}{\texttt{"If Amazon place\_order call fails due to an invalid promo code, use remove\_promo\_code\_from\_cart tool to remove it."}}
\trow{Agent:}{\textit{I need to remove the invalid promo code first. Let me double-check the signature:}}
\trow{}{\texttt{show\_api\_doc(app\_name="amazon",\allowbreak{} api\_name="remove\_promo\_code\_from\_cart")}}
\trow{Env:}{Description: Removes a promo code from the cart.}
\trow{}{Parameters: access\_token (str) ...}
\trow{Agent:}{\texttt{remove\_promo\_code\_from\_cart(\allowbreak{}access\_token="0ac87b...")}}
\trow{Env:}{\texttt{"Removed the promo code from the cart."}}
\trow{Agent:}{\texttt{place\_order(payment\_card\_id=64, address\_id=29,\allowbreak{} access\_token="0ac87b...")}}
\trow{Env:}{"Order placed successfully. Debited a total of 2690.76 from your card."}
\trow{Agent:}{\texttt{complete\_task(status="success")}}
\end{transcript}
\caption{Example of \textit{warning}.}
\label{fig:warning-example}
\end{figure}

\subsection{Examples of constraint application}

The warning and repair modes show complementary strengths.
Warning mode catches failures that only show in the outcome, such as an expired payment card or an invalid promo code, which no check before the call can detect.
On the other hand, repair mode fixes mistakes visible in the call itself. For instance, GPT-5.4-mini often mistypes long access tokens (82\% of its failed calls are authentication errors), and repair removes 68\% of these errors.
%
%
In this section, we show two qualitative examples of how the warning and repair mechanisms guide the agent towards correct tool calling.

\paragraph{Warning.} (AppWorld task \texttt{77bcb81\_3}, Sonnet 5) In this task, the user asked the agent to place an order for every item in the Amazon cart and wish list. After moving the wish-list items into the cart, the agent attempted checkout, but an expired promotion code was registered to the cart and invalidated the order.
The constraint identified the failure case by the returned error message of the \textsc{PlaceOrder} tool, providing the detailed reason for the failure and guiding the agent to the correct tool. The excerpt in \autoref{fig:warning-example} preserves the recorded inputs and outputs while redacting the access token. The agent directly called the correct tool for removing the promotion code, retried the original order, and passed the task. 

\begin{figure}[]
\small
\begin{transcript}{Repair:}
\trow{Agent:}{\texttt{gmail\_login(id="", password="")}}
\trow{Env:}{Access token: "378c1a..."}
\trow{Agent:}{\texttt{filesystem\_login(id="", password="")}}
\trow{Env:}{Access token: "f01829..."}
\trow{}{...}
\trow{Agent:}{\texttt{show\_file(file\_path="\textasciitilde/documents/}}
\trow{}{\texttt{\ \ personal/bradley\_trip\_expenses.pdf",}}
\trow{}{\texttt{\ \ access\_token="378c1a...")}}
\scarow{Repair:}{\texttt{access\_token: "378c1a..." $\mapsto$ "f01829..."}}
\trow{Env:}{File metadata and contents returned successfully.}
\trow{Agent:}{\texttt{upload\_attachments\_to\_draft(draft\_id=518,}}
\trow{}{\texttt{\ \ attachment\_file\_paths=[...], access\_token="378c1a...",}}
\trow{}{\texttt{\ \ file\_system\_access\_token="378c1a...")}}
\scarow{Repair:}{\texttt{file\_system\_access\_token:}}
\scarow{}{\texttt{\ \ "378c1a..." $\mapsto$ "f01829..."}}
\trow{Env:}{The local PDF was attached to draft 518.}
\trow{Agent:}{\texttt{send\_email\_from\_draft(draft\_id=518, access\_token="378c1a...",}}
\trow{}{\texttt{\ \ file\_system\_access\_token="378c1a...")}}
\scarow{Repair:}{\texttt{file\_system\_access\_token:}}
\scarow{}{\texttt{\ \ "378c1a..." $\mapsto$ "f01829..."}}
\trow{Env:}{Draft 518 was sent successfully.}
\trow{}{...}
\trow{Agent:}{\texttt{complete\_task(status="success")}}
\end{transcript}
\caption{Example of \textit{repair}.}
\label{fig:repair-example}
\end{figure}

\paragraph{Repair.} (AppWorld task \texttt{4ac4a8d\_3}, GPT-5.4-mini) In this task, the user asked the agent to forward an email thread to a given contact and attach the given file. When searching for the file via FileSystem APIs, the agent repeatedly supplied the email application's credentials instead, which should have failed the tool execution. The plain tool-calling baseline repeated the tool call with the invalid token, not realizing the mistake, and eventually exhausted the turn limit.
\autoref{fig:repair-example} shows
SCA with repair mode automatically corrects the invalid credential token to a correct one, as the correct token for the \texttt{show\_file} tool uniquely exists in the agent's previous tool-calling history. Each substitution changed only one argument leaf and preserved the remainder of the agent's call. After automatic repair, the agent proceeds to attach the file and forward the email, completing the task. 

%% file: conclusion.tex
\section{Conclusion}
    
SCA mines tool-call constraints in formal logic from tool definitions and corrects them against agent traces. Checked by a solver at runtime, these constraints detect faulty calls and raise task success while cutting failed calls and tokens through warnings and repair. 
We demonstrate that these improvements generalize to realistic settings without explicit domain policies or informative error messages, highlighting the potential of our approach for building more controllable and trustworthy agents.